\documentclass{article}

\usepackage[preprint]{neurips_2025}

\usepackage[utf8]{inputenc} 
\usepackage[T1]{fontenc}    
\usepackage{hyperref}       
\usepackage{url}            
\usepackage{booktabs}       
\usepackage{amsfonts}       
\usepackage{nicefrac}       
\usepackage{microtype}      
\usepackage{enumitem}
\usepackage{multirow}
\usepackage{graphicx}
\usepackage{amsmath} 
\usepackage{arydshln}
\usepackage[normalem]{ulem}
\useunder{\uline}{\ul}{}
\usepackage[misc]{ifsym}
\usepackage[table]{xcolor}
\usepackage{algorithm}
\usepackage{algpseudocode}

\title{Fre-Res: Frequency-Residual Video Token Compression for Efficient Video MLLMs}

\author{Yigui Feng\textsuperscript{1},
        Qinglin Wang\textsuperscript{1*},
        Yang Liu\textsuperscript{2},
        and Jie Liu\textsuperscript{1}%
\thanks{\textsuperscript{1}All authors with superscript 1 are with the College of Computer Science, National University of Defense Technology, Deya Road 109, Changsha, Hunan 410073, China.}%
\thanks{\textsuperscript{2}Yang Liu is with the Shien-Ming Wu School of Intelligent Engineering, South China University of Technology, Xingye Road 777, Guangzhou, Guangdong 511442, China.}%
\thanks{*Corresponding author:Jie Liu, Qinglin Wang (e-mail: wangqinglin@nudt.edu.cn).}%
}

\begin{document}

\maketitle

\begin{abstract}
Video MLLMs face a persistent tension between spatial fidelity and temporal coverage: preserving fine-grained visual details requires many spatial tokens, while capturing short-lived events requires dense temporal sampling.
We propose \textbf{Fre-Res}, a budget-adaptive dual-track video-token compression framework that separates these two forms of evidence.
Fre-Res preserves sparse high-fidelity spatial anchors and represents dense temporal evolution through compact residual-frequency tokens.
Specifically, it applies temporal 1D-DCT to inter-frame residual trajectories in vision-latent space, where we observe strong low-frequency concentration.
To align frequency-domain dynamics with native visual embeddings, Fre-Res introduces a Spatial-Guided Absorber that injects temporal residual information into spatially corresponding anchor tokens.
Across fine-grained short-video and long-video reasoning benchmarks, Fre-Res achieves a favorable accuracy--efficiency trade-off, matching or approaching full-token performance while substantially reducing visual-token length.
Extensive ablations further show that temporal-frequency residuals preserve causal transition cues, while spatial anchors remain essential for fine-grained object and layout reasoning.
\end{abstract}

\section{Introduction}
\label{sec:intro}

Large Language Models (LLMs) have shown strong capability in modeling long symbolic sequences~\citep{grattafiori2024llama3}, and recent multimodal extensions have pushed this capability toward video understanding~\citep{li2024llavanextstrong}.
However, long-video MLLMs face a fundamental efficiency challenge.
A video contains dense and highly redundant spatiotemporal observations, while current MLLMs typically convert sampled frames into long sequences of visual embeddings.
As the number of frames increases, this leads to quadratic prefill attention cost and linearly growing KV-cache memory, making dense long-video processing difficult to scale~\citep{li2024snapkv,shao2025tokens}.

The challenge is not only the total number of visual tokens, but also the different roles played by spatial and temporal information.
Fine-grained video understanding requires \textit{spatial fidelity} to recognize objects, attributes, positions, counts, and local configurations.
At the same time, temporal reasoning requires sufficient \textit{temporal coverage} to capture actions, transitions, and causal events.
Under a limited visual-token budget, these two requirements often compete with each other.
Reducing spatial tokens may remove small objects or precise layouts, while sparse frame sampling may miss short-lived events.
Therefore, effective video-token compression should not treat all visual tokens as a homogeneous sequence.
Instead, it should preserve spatial evidence and temporal dynamics in a structured and complementary manner~\citep{shao2025holitom}.

\begin{figure}[t]
    \centering
    \includegraphics[width=\textwidth]{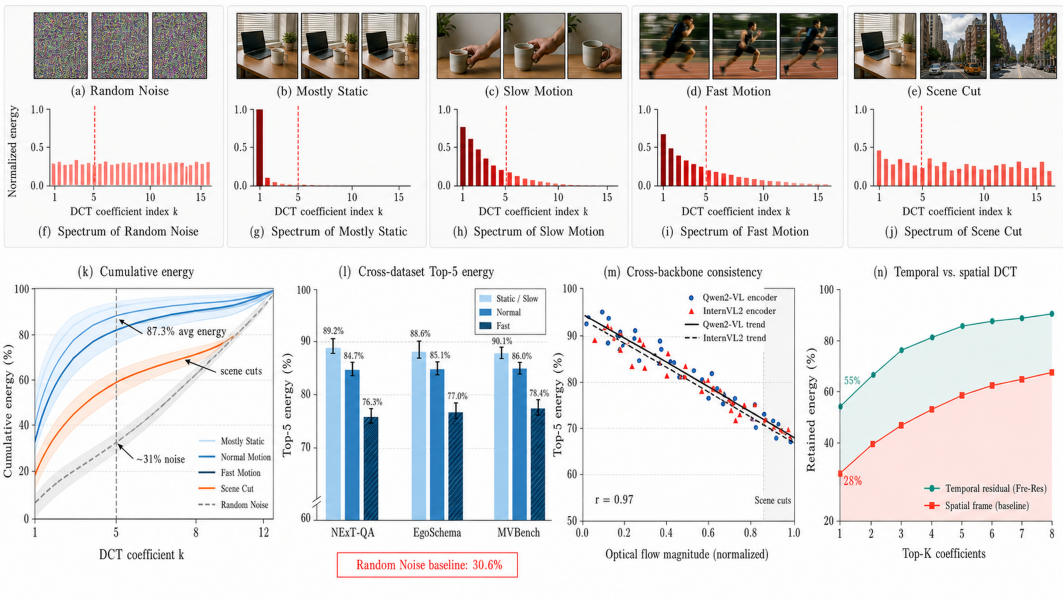}
\caption{\textbf{Temporal-frequency energy concentration in vision-latent residuals.}
\textbf{(a--e)} Example frame sequences: random noise, mostly static scene, slow motion,
fast motion, and scene cut.
\textbf{(f--j)} Corresponding temporal 1D-DCT energy spectra of latent residual trajectories.
Real video residuals concentrate energy in low-frequency components, while random noise
distributes energy uniformly. Concentration weakens progressively from static to fast
motion, and is further reduced at scene cuts---motivating our event-aware I-frame selection.
\textbf{(k)} Cumulative energy curves sampled from 400 clips across NExT-QA, EgoSchema,
and MVBench (Qwen2-VL encoder): the first 5 DCT coefficients capture \textbf{87.3\%\,$\pm$\,4.2\%}
of total residual energy on average, versus only $\sim$31\% for random noise.
Scene cuts (orange) exhibit substantially lower concentration than normal motion,
confirming that I-frame boundaries should be event-aware rather than uniformly spaced.
\textbf{(l)} Cross-dataset Top-5 energy ratios remain consistently high (76--90\%) across
all three benchmarks and motion types, well above the random noise baseline of 30.6\%,
demonstrating that low-frequency concentration is a robust, dataset-independent property.
\textbf{(m)} Scatter of Top-5 energy vs.\ optical flow magnitude for two visual encoders
(Qwen2-VL and InternVL2) yields a Pearson correlation of $r{=}0.97$ between the two
trend lines, confirming that this concentration is a structural property of natural video
rather than an encoder-specific artifact.
\textbf{(n)} Temporal residual DCT (Fre-Res) retains substantially more energy per
coefficient than spatial frame DCT: at Top-1, temporal residual captures 55\% versus
28\% for spatial, validating the advantage of applying frequency compression to
inter-frame residual trajectories rather than individual frames.}
    \label{fig:teaser}
\end{figure}

Fre-Res is motivated by this observation.
Classical video codecs exploit inter-frame prediction and transform coding to separate reference information from residual dynamics.
Inspired by this principle, we move residual-based compression from pixel space to the latent space of vision encoders.
Our empirical analysis shows that inter-frame residual trajectories in vision-latent space exhibit strong low-frequency concentration under temporal 1D Discrete Cosine Transform (1D-DCT), as illustrated in Figure~\ref{fig:teaser}.
Figure~\ref{fig:teaser} provides the key motivation for our temporal branch.
Random noise exhibits scattered spectral energy after temporal 1D-DCT, whereas real video residuals show much stronger low-frequency concentration.
Moreover, the spectrum changes with motion intensity: mostly static scenes concentrate energy near the DC component, while faster or more interactive motions spread energy over a broader low-frequency range.
This observation suggests that temporal residuals exhibit structured, compressible energy 
patterns in latent space, containing motion evidence that can be represented compactly 
in the frequency domain.
Therefore, temporal changes between neighboring frames can often be represented compactly by a small number of low-frequency residual coefficients.

Based on this observation, we propose \textbf{Fre-Res}, a budget-adaptive dual-track video-token compression framework.
The first track preserves high-fidelity spatial evidence through sparse raw anchors.
These anchors retain local spatial coverage and provide object-level and layout-level information.
The second track represents dense temporal evolution through compact residual-frequency tokens.
Specifically, Fre-Res computes latent residuals between non-anchor frames and their nearby 
anchor frames, applies temporal 1D-DCT along the residual trajectories, and retains the 
first $K$ low-frequency coefficients as temporal P-Tokens, where $K$ is determined by 
the remaining token budget $B_\text{freq}$.
In this way, spatial details and temporal dynamics are compressed through different mechanisms rather than being forced to compete under a single pruning rule.

A remaining challenge is that frequency-domain residual tokens are not native visual tokens and may be difficult for a frozen LLM to interpret directly~\citep{wang2025fourier}.
To address this semantic gap, Fre-Res introduces a lightweight \textbf{Spatial-Guided Absorber}.
The absorber uses local masked cross-attention to inject temporal-frequency residual information into spatially corresponding anchor tokens.
This produces motion-aware spatial tokens while avoiding explicit inverse reconstruction of dense visual grids.
We further keep compact frequency-summary tokens to retain global temporal signatures.

Fre-Res can be instantiated under different visual-token budgets.
For strict short-video comparisons, we use a compact budget-matched setting to compare fairly with spatial-compression baselines.
For longer videos, Fre-Res increases the temporal aggregation range to keep the sequence length practical while preserving sparse spatial anchors and compact temporal-frequency evidence.
Across fine-grained short-video and long-video reasoning benchmarks, Fre-Res provides a favorable accuracy--efficiency trade-off compared with token pruning~\citep{chen2026evoprune}, token merging~\citep{lee2024dtem}, and spatial frequency compression~\citep{li2023dct} baselines.

In summary, our contributions are three-fold:
\begin{itemize}
    \item \textbf{Latent temporal-frequency observation.}
    We empirically show that inter-frame residuals in vision-latent space exhibit strong low-frequency concentration under temporal 1D-DCT, revealing a compressible structure that is not captured by frame-wise spatial token pruning.

    \item \textbf{Budget-adaptive dual-track compression.}
    We propose Fre-Res, a dual-track video-token compression framework that preserves high-fidelity spatial anchors while representing dense temporal evolution through compact residual-frequency tokens.

    \item \textbf{Frequency-aware token fusion.}
    We introduce a Spatial-Guided Absorber that aligns frequency-domain residual dynamics with spatial visual tokens, enabling efficient video reasoning under reduced visual-token budgets.
\end{itemize}

\section{Related Work}
\label{sec:related_work}

\subsection{Long-Context Video MLLMs}

Recent progress in multimodal large language models has enabled increasingly capable video understanding systems.
Representative models such as Qwen2-VL~\citep{wang2024qwen2vl}, LongVA~\citep{zhang2024longcontext}, and LLaVA-Video extend image-language models to video inputs through improved visual tokenization, dynamic resolution processing, multimodal positional encoding, and large-scale video instruction tuning.
These models demonstrate that longer temporal context can substantially improve video reasoning, especially for tasks involving actions, events, and causal relationships.

However, directly scaling the number of visual tokens remains expensive.
When dense frames are converted into long visual-token sequences, the prefill attention cost grows quadratically with sequence length, while the KV-cache memory grows linearly.
As a result, processing high-frame-rate or long-duration videos can become impractical even for strong MLLM backbones.
This motivates efficient video-token compression methods that reduce redundant visual evidence while preserving the information required for spatial and temporal reasoning.

\subsection{Visual Token Compression and Pruning}

A growing line of work reduces inference cost by pruning or merging visual tokens.
Attention-based methods, such as FastV~\citep{chen2024imagehalf} and DyCoke~\citep{tao2025dycoke}, estimate token importance from the model's internal attention patterns and discard tokens with lower contribution.
Other methods exploit redundancy among visual embeddings through clustering, similarity-based merging, or diversity-aware selection, such as LLaVA-PruMerge~\citep{shang2025llavaprumerge}, CDPruner~\citep{zhang2025conditionaldiversity}, and FrameFusion~\citep{fu2025framefusion}.
These approaches are attractive because they can often be applied to existing MLLMs with limited architectural changes.
Recent work has further argued that token reduction should be viewed not merely as a post-hoc efficiency optimization, but as a task-aware design principle that preserves semantic integrity and supports multimodal integration~\citep{kong2025tokenreduction}.

From this perspective, a key limitation of many existing pruning or merging methods is that they treat visual tokens as a relatively homogeneous sequence.
For videos, this can be limiting because spatial fidelity and temporal coverage play different roles.
Removing spatial tokens may harm object identity, position, counting, and pose reasoning, while sparse frame selection may miss short-lived transitions or causal events.
Fre-Res follows this task-aware perspective by explicitly separating spatial anchors from temporal residual dynamics.
Instead of applying a single pruning rule to all tokens, it preserves high-fidelity spatial evidence through raw anchors and compresses inter-frame dynamics through temporal-frequency residuals.

\subsection{Frequency-based Compression and Codec Priors}

Frequency-domain transformations have long been used for efficient signal representation.
In neural sequence modeling, spectral mixing or frequency-domain compression has been explored in architectures such as FNet~\citep{leethorp2022fnet} and FreqKV~\citep{kai2026freqkv}.
In vision and multimodal learning, recent methods such as Fourier-VLM~\citep{wang2025fourier} show that visual embeddings often contain spatial low-frequency redundancy, enabling token reduction through frequency-domain processing.

Most existing frequency-based visual compression methods focus on intra-frame spatial redundancy.
This is effective for images or individual frames, but it does not directly model inter-frame temporal redundancy, which is central to video understanding.
Moreover, frequency-domain features may not be directly aligned with the visual-token manifold expected by frozen LLMs, creating a semantic gap between compressed spectral representations and native visual embeddings.

Fre-Res builds on the general principle of transform-based compression but applies it to temporal residual trajectories in vision-latent space.
Inspired by classical video codecs, which separate reference information from residual dynamics through inter-frame prediction and transform coding, Fre-Res preserves sparse spatial anchors and applies temporal 1D-DCT to latent residuals between frames.
To make these residual-frequency tokens usable by the LLM, Fre-Res further introduces a Spatial-Guided Absorber that aligns temporal-frequency dynamics with spatial anchor tokens before language decoding.

\section{Methodology}
\label{sec:method}

\begin{figure*}[t]
    \centering
    \includegraphics[width=\textwidth]{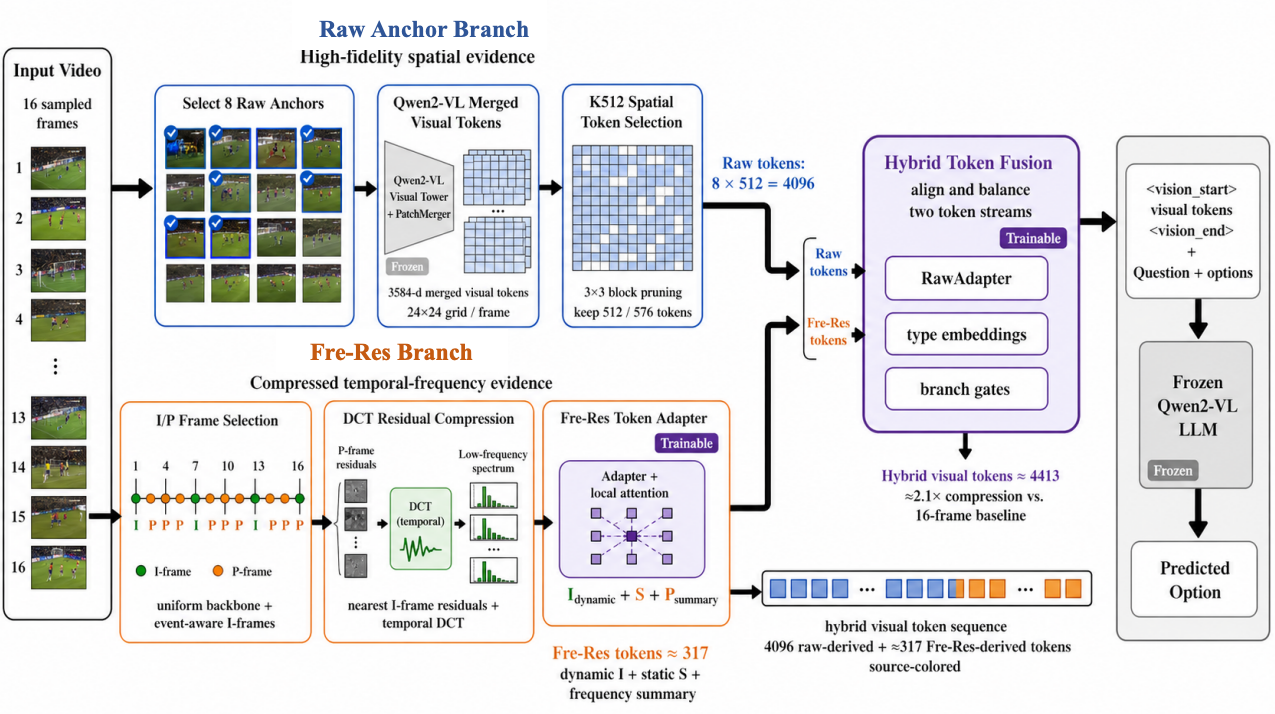}
    \caption{\textbf{The Dual-Branch Architecture of Fre-Res.}
The framework is illustrated using a standard 16-frame configuration as an example.
\textbf{Raw Anchor Branch:} Selects sparse keyframes (e.g., 8 anchors) and applies parameter-free $3\times3$ block pruning to preserve 512 out of 576 tokens per frame, retaining high-fidelity spatial evidence.
\textbf{Fre-Res Branch:} Generates compressed temporal-frequency evidence. Temporal 1D-DCT extracts low-frequency residual components, which are then organized into dynamic anchors ($I_{\text{dynamic}}$), static tokens ($S$), and global frequency summaries ($P_{\text{summary}}$).
\textbf{Hybrid Token Fusion:} A unified module aligns and balances the two streams via a RawAdapter, type embeddings, and branch gates.
In the illustrated 16-frame setting, this instantiation yields $\approx 4413$ visual tokens, corresponding to about $2.1\times$ compression.
In practice, Fre-Res is budget-adaptive: the number of anchors, the spatial token-retention ratio, and the temporal aggregation range can be adjusted according to the input duration and the target visual-token budget.}
    \label{fig:architecture}
\end{figure*}

A robust video-token compression mechanism should exploit the structure of video signals rather than relying solely on unstructured token dropping.
In this section, we formulate the tension between spatial fidelity and temporal coverage, and present \textbf{Fre-Res} as a structured dual-track design motivated by this tension.
The key idea is to preserve high-fidelity spatial evidence through sparse anchors while representing dense temporal evolution through compact residual-frequency tokens.

\subsection{Problem Formulation: Spatial Fidelity vs. Temporal Coverage}
\label{subsec:formulation}

Given a video sequence of $T$ frames, each frame is converted by the visual encoder into $N$ visual embeddings.
Flattening all visual embeddings into a single sequence yields $TN$ visual tokens, leading to a self-attention prefill cost of $\mathcal{O}((TN)^2)$ and a KV-cache memory cost that grows linearly with $TN$~\citep{munkhdalai2024infiniattention}.
Under a limited visual-token budget, compression is therefore necessary.

However, video understanding requires two different types of evidence.
Static structures require \textit{high spatial fidelity} to preserve object identity, attributes, positions, counts, and local configurations.
Dynamic events require \textit{temporal coverage} to capture actions, transitions, and causal relations.
Treating all visual tokens as a homogeneous sequence can make these two requirements compete with each other:
aggressive spatial pruning may remove fine-grained visual details, while sparse temporal sampling may miss short-lived events.
This motivates a \textit{divide-and-conquer} formulation: we separate the signal into a \textbf{Raw Anchor Branch} for high-fidelity spatial evidence and a \textbf{Fre-Res Branch} for compact temporal dynamics.

\subsection{Raw Anchor Branch: High-Fidelity Spatial Evidence}
\label{subsec:raw_branch}
To satisfy spatial fidelity, the Raw Anchor Branch extracts a sparse set of $M$ frames (e.g., 8 raw anchors) from the 16 sampled frames.
These frames are processed by the visual tower (e.g., Qwen2-VL), outputting merged visual tokens forming a $24 \times 24$ spatial grid.

Global attention-based pruning may remove spatially contiguous regions, which can weaken the global spatial layout and harm fine-grained reasoning.
Instead, we implement a \textit{parameter-free $3\times3$ block pruning} selection.
We partition the $24 \times 24$ grid into local neighborhoods ($3 \times 3$ blocks).
By deterministically discarding the single token with the lowest energy within each block, we preserve exactly 512 out of 576 tokens per frame.
This preserves local spatial coverage and avoids deleting entire macroscopic regions, which is important for object identity, layout, counting, and position-sensitive reasoning~\citep{xu2025adallava}.

\subsection{Fre-Res Branch: Compressed Temporal-Frequency Evidence}
\label{subsec:fre_res_branch}

The Fre-Res Branch captures the \textit{inter-frame dynamics} of the full 16-frame sequence
to satisfy dense temporal coverage.

\textbf{I/P Frame Selection.}
We establish a uniform backbone of evenly-spaced I-frames supplemented by event-aware
I-frames: a frame $t$ is promoted to an I-frame if its cosine distance to the previous
frame in latent space exceeds a threshold $\tau{=}0.3$, otherwise it is treated as a
P-frame.
This design ensures that abrupt scene transitions---where low-frequency concentration
is weaker, as shown in Figure~\ref{fig:teaser}(e)(j)(k)---are handled by inserting
a new I-frame anchor rather than relying on residual compression across the cut.
The temporal information of each P-frame is then encapsulated in the residual
$\Delta F_t = F_t - F_{a(t)}$, where $a(t)$ denotes the nearest preceding I-frame.

\textbf{Temporal 1D-DCT compression.}
Because physical actions evolve continuously over short temporal windows, the corresponding
latent residual trajectories vary smoothly, making the frequency domain a natural
representation for compact temporal modeling.
We apply the Type-II 1D Discrete Cosine Transform (1D-DCT) along the temporal axis and
retain the first $K$ low-frequency coefficients as P-Tokens, where $K$ is determined
by the remaining token budget $B_\text{freq}$ (see Section~\ref{subsec:fusion}).
The validity of this compression step is directly supported by the dataset-level analysis
in Figure~\ref{fig:teaser}(k--n): across 400 clips sampled from NExT-QA, EgoSchema,
and MVBench, the first 5 DCT coefficients capture $87.3\%\,{\pm}\,4.2\%$ of total
residual energy under both Qwen2-VL and InternVL2 encoders ($r{=}0.97$ between backbones),
confirming that low-frequency concentration is a robust structural property of natural
video rather than an anecdotal observation.

\subsection{Fre-Res Token Adapter: Bridging the Semantic Gap}
\label{subsec:absorber}
Directly feeding frequency-domain P-Tokens to a frozen LLM can create a semantic mismatch, since these tokens do not follow the same distribution as native visual embeddings.
To reduce this gap, the Fre-Res Token Adapter aligns temporal-frequency dynamics with spatial anchor tokens through a \textbf{Spatial-Guided Cross-Attention} mechanism.
The static I-Tokens act as queries and locally attend to their corresponding P-Tokens under a spatial neighborhood mask, producing motion-aware dynamic anchors ($I_{\text{dynamic}}$)~\citep{makitalo2024bridging}.

To ensure comprehensive evidence coverage, the Fre-Res Branch ultimately outputs three sets of tokens: (1) \textbf{$I_{\text{dynamic}}$} for local motion-infused anchors; (2) \textbf{$S$} for static background patches; and (3) \textbf{$P_{\text{summary}}$}, which is the globally pooled low-frequency spectrum serving as a global frequency summary token, computed by averaging the retained low-frequency DCT coefficients across all P-frames in the GoP. This module yields an ultra-compact sequence of $\approx 317$ tokens in the compact 16-frame instantiation.
Concretely, after temporal grouping and spatial pooling, each temporal group produces a small set of candidate P-Tokens from the retained low-frequency residual coefficients.
In our compact setting, we use 8 temporal groups, retain $K{=}3$ low-frequency coefficients, and pool residuals onto a $4{\times}4$ spatial grid, producing up to
$8 \times 3 \times 16 = 384$ candidate P-Tokens before energy-based filtering.
After removing near-zero-energy positions, the number of dynamic P-Tokens is reduced to approximately 285 on average.
Together with 8 global frequency-summary tokens ($P_{\text{summary}}$, one per temporal group) and approximately 24 static background tokens ($S$), the Fre-Res branch yields about
$285 + 8 + 24 \approx 317$ tokens.

\subsection{Hybrid Token Fusion: Alignment and Balancing}
\label{subsec:fusion}
The final step is to integrate the spatial-anchor stream and the condensed temporal-frequency stream into a cohesive sequence for the frozen LLM. Because these two streams originate from different representational manifolds, we introduce a \textbf{Hybrid Token Fusion} module. Specifically, the raw-anchor tokens are first passed through a lightweight RawAdapter to align their distribution with the multimodal fusion space.
We then add type embeddings to indicate whether each token comes from the raw spatial branch or the Fre-Res temporal branch.
Finally, trainable branch gates are used to balance the magnitudes of the two streams, preventing the high-volume raw-anchor tokens from overwhelming the compact but causally informative temporal-frequency tokens.
The fused visual sequence is then injected into the frozen LLM as visual embeddings.

Crucially, the dual-track formulation is naturally budget-adaptive.
Figure~\ref{fig:architecture} illustrates one standard 16-frame instantiation, where selecting $M=8$ anchors and applying the $8/9$ local energy filter yields 4096 spatial tokens.
Combined with approximately 317 compact Fre-Res tokens (derived in Section~\ref{subsec:absorber}), the final sequence contains about 4413 visual tokens, corresponding to roughly $2.1\times$ compression relative to the full-token baseline in this example setting.

More generally, Fre-Res can be instantiated under different token budgets by adjusting the anchor count, the spatial token-retention ratio, and the temporal aggregation range.
For strict short-video comparisons such as MVBench, we use a budget-matched compact instantiation with a reduced anchor budget so that the total number of visual tokens is comparable to, or smaller than, spatial-compression baselines.
For temporal and long-video reasoning, the temporal aggregation range is enlarged to cover longer inputs while keeping the overall sequence length bounded.
For example, in our dense 30-FPS long-video profiling setting, Fre-Res reduces a sequence of more than one million visual tokens to about 46K tokens, corresponding to more than $20\times$ compression while retaining sparse spatial anchors and compact temporal-frequency evidence~\citep{wang2025lvbench}.

\textbf{Budget allocation rule.}
Fre-Res uses a hierarchical rather than end-to-end learned budget allocation.
Given a target visual-token budget $B$, we first reserve tokens for the raw spatial anchors,
$B_{\mathrm{spatial}} = M \cdot K_{\mathrm{raw}}$,
where $M$ is the number of selected anchor frames and $K_{\mathrm{raw}}$ is the number of retained raw tokens per anchor.
The remaining budget is assigned to the temporal Fre-Res branch after reserving summary and static tokens.
The number of retained DCT coefficients is then chosen according to the available temporal-branch budget and capped by the effective temporal length of each temporal group.
For compact short-video settings, we reduce $M$ and $K_{\mathrm{raw}}$ to match the target budget; for longer videos, we increase the temporal aggregation range and adjust anchor density to keep the sequence length bounded. Hyperparameter sensitivity is reported in Appendix~\ref{app:sensitivity}.
\section{Experiments}
\label{sec:experiments}

Our experiments are designed to validate whether Fre-Res provides a better accuracy--efficiency trade-off than spatial frequency compression under comparable or stricter visual-token budgets. 
Rather than solely pursuing leaderboard performance, we focus on isolating the empirical effects of temporal-frequency decoupling, spatial anchor preservation, and hybrid token fusion.

Specifically, we investigate four questions:
\textbf{RQ1 (Fine-Grained Spatial Reasoning):} Under the same backbone and a no-larger visual-token budget, does Fre-Res preserve fine-grained short-video understanding better than spatial frequency compression?
\textbf{RQ2 (Temporal and Long-Video Reasoning):} Can Fre-Res preserve temporal and causal evidence on NExT-QA, LongVideoBench, and EgoSchema while substantially reducing the number of visual tokens?
\textbf{RQ3 (Practical Efficiency):} Does the visual-token reduction of Fre-Res translate into real hardware savings in memory usage and prefill latency on dense long-video inputs?
\textbf{RQ4 (Component Analysis):} How do the spatial anchors, temporal-frequency residuals, frequency summaries, and Spatial-Guided Absorber individually contribute to the final performance?

\subsection{Experimental Setup}
\label{subsec:exp_setup}

\textbf{Backbones.}
We evaluate Fre-Res under two groups of backbone settings.
For fine-grained short-video evaluation on MVBench, we use InternVL2~\citep{opengvlab2024internvl2}, MiniCPM-v2.6~\citep{yao2024minicpmv}, InternVL2.5~\citep{chen2024internvl25}, \textbf{Qwen2-VL-2B} and \textbf{Qwen3-VL-4B}~\citep{bai2025qwen3vl}, which allow controlled comparisons under relatively compact visual-token budgets.
For temporal and long-video reasoning benchmarks, including NExT-QA~\citep{xiao2021nextqa}, LongVideoBench~\citep{wu2024longvideobench}, and EgoSchema~\citep{mangalam2023egoschema}, we use stronger \textbf{Qwen2-VL-7B} and \textbf{Qwen3-VL-8B} backbones to evaluate whether Fre-Res scales to more capable video MLLMs.
Hardware profiling is conducted with \textbf{Qwen3-VL-8B} under the dense long-video setting.
All controlled comparisons are performed within the same backbone family and evaluation protocol.

\textbf{Benchmarks.}
We use \textbf{MVBench}~\citep{li2024mvbench} as the primary benchmark for fine-grained short-video understanding, since it covers action recognition, object reasoning, spatial position, counting, pose, and causal cognition.
We further evaluate on \textbf{NExT-QA}, \textbf{LongVideoBench}, and \textbf{EgoSchema} using accuracy as the primary metric to test long-horizon temporal and causal reasoning.

\textbf{Evaluation protocol.}
For MVBench, all models are evaluated using the same \texttt{lmms-eval} protocol, including the prompt format, decoding configuration, and answer extraction rule.
For NExT-QA, LongVideoBench, and EgoSchema, we report standard accuracy following their official evaluation settings.
Unless otherwise specified, \#V denotes the number of visual embeddings injected into the LLM after compression/projection, excluding text tokens and special prompt tokens.

\textbf{Budgeted instantiations.}
Fre-Res is a budget-adaptive framework rather than a single fixed-token configuration.
Accordingly, different experiments instantiate the same dual-track design under different input lengths and token budgets.
For short-video budget-matched comparisons, such as MVBench in Table~\ref{tab:mvbench}, we use a compact instantiation whose visual-token count is constrained to be no larger than the corresponding spatial-compression baseline.
For temporal and long-video reasoning in Table~\ref{tab:long_video}, we use a longer-context instantiation with a larger visual-token budget to preserve event-level temporal evidence.
For dense-video hardware profiling in Table~\ref{tab:hardware}, we use the long-video instantiation that enlarges the temporal aggregation range to keep the total sequence length practical for high-frame-rate inputs.
All reported token counts therefore correspond to the actual number of visual embeddings injected into the LLM under each experimental budget.

\textbf{Baseline fairness.}
For controlled comparisons, all baselines are evaluated under the same backbone, prompt format, decoding configuration, and answer extraction rule.
We use the official hyperparameters of each baseline whenever available, and tune only the primary budget-control parameter to match the target visual-token count.
Specifically, we adjust the attention retention threshold for DyCoke, the similarity threshold for CDPruner, and the frequency truncation level for Fourier-VLM.
No task-specific re-training is applied to the compression baselines.
To reduce sensitivity to a single operating point, we additionally check nearby token budgets and report the best or representative matched-budget result under the same backbone.
Detailed budget-sweep results are provided in Appendix~\ref{app:baseline_fairness}.

\subsection{Fine-Grained Spatial Reasoning on MVBench (RQ1)}
\label{subsec:rq1_mvbench}

MVBench is used as our primary benchmark for fine-grained video understanding, as it covers a broad range of abilities, including action recognition, object perception, spatial position, counting, pose, character understanding, and causal cognition. 
We use this benchmark to examine whether Fre-Res offers a better accuracy--efficiency trade-off than spatial frequency compression.

Table~\ref{tab:mvbench} contains two types of comparisons. 
The first block reports representative MLLM baselines under their default visual-token budgets, providing a reference point for the overall MVBench scale. 
These models differ in architecture, training data, visual encoders, and tokenization strategies, so they are not intended as controlled compression baselines. 
The lower blocks are the main comparison of interest: for Qwen2-VL and Qwen3-VL, we compare the full-token vanilla model, Fourier-based spatial compression, and Fre-Res under the same backbone. 
To make the comparison conservative, Fre-Res is constrained to use no more visual tokens than the corresponding Fourier baseline.

\begin{table}[t]
\centering
\caption{\textbf{MVBench comparison under stricter visual-token budgets.}
All MVBench results in this table are evaluated using the same \texttt{lmms-eval} protocol.
The first block reports representative MLLM baselines under their default visual-token budgets, while the lower blocks provide controlled compression comparisons under the same backbone.
For each controlled backbone, \textbf{Fre-Res-Compact} denotes the compact budget-matched instantiation of Fre-Res whose visual-token count is no larger than the corresponding Fourier-based spatial compression baseline.
Under this stricter budget, Fre-Res-Compact preserves overall accuracy more effectively than spatial compression, especially on Qwen-family backbones.
The category-level results show that Fre-Res-Compact better maintains Action, Position, Count, and Scene performance, while fine-grained Pose remains more sensitive to compression.
Avg. follows the original MVBench task-level aggregation, and \#V denotes the number of visual embeddings injected into the LLM after compression/projection, excluding text and special prompt tokens.}
\label{tab:mvbench}
\resizebox{\textwidth}{!}{
\begin{tabular}{lcc | ccccccccc | c}
\toprule
\textbf{Model} & \textbf{Size} & \textbf{\#V} & \textbf{Action} & \textbf{Object} & \textbf{Position} & \textbf{Scene} & \textbf{Count} & \textbf{Attribute} & \textbf{Pose} & \textbf{Character} & \textbf{Cognition} & \textbf{Avg.} \\
\midrule
\multicolumn{13}{l}{\textit{\textbf{Reference MLLM baselines}}} \\
InternVL2 & 2B & 4096 & 48.72 & 69.81 & 56.41 & 87.92 & 42.11 & 49.63 & 61.07 & 63.24 & 36.52 & 58.21 \\
MiniCPM-V 2.6 & 3B & 1024 & 49.03 & 73.57 & 55.92 & 88.74 & 50.21 & 55.34 & 64.21 & 70.48 & 38.41 & 60.31 \\
InternVL2.5 & 2B & 4096 & 51.94 & 73.82 & 59.73 & 90.87 & 46.54 & 53.48 & 64.82 & 67.11 & 40.21 & 61.28 \\
\midrule
\multicolumn{13}{l}{\textit{\textbf{Backbone: Qwen2-VL}}} \\
Vanilla (Full Token Bound) & 2B & 3193 & \textbf{68.52} & \textbf{65.74} & \textbf{44.86} & \textbf{90.51} & \textbf{60.83} & \textbf{65.37} & \textbf{53.58} & \textbf{59.04} & 47.85 & \textbf{61.43} \\
Fourier-Qwen-2 (Spatial) & 2B & 1328 & 67.75 & 63.32 & 40.05 & 88.54 & 55.51 & 64.82 & 51.06 & 57.08 & \textbf{50.21} & 59.86 \\
\textit{$\Delta$ vs. Vanilla} & & \textit{-58.4\%} & \textit{-0.77} & \textit{-2.42} & \textit{-4.81} & \textit{-1.97} & \textit{-5.32} & \textit{-0.55} & \textit{-2.52} & \textit{-1.96} & \textit{+2.36} & \textit{-1.57} \\
\rowcolor{blue!5} \textbf{Fre-Res-Compact (Ours)} & 2B & \textbf{1280} & 68.87 & 65.06 & 44.15 & 90.22 & 59.51 & 65.04 & 49.52 & 58.26 & 45.53 & 61.05 \\
\rowcolor{blue!5} \textit{$\Delta$ vs. Vanilla} & & \textbf{\textit{-59.9\%}} & \textbf{\textit{+0.35}} & \textbf{\textit{-0.68}} & \textbf{\textit{-0.71}} & \textbf{\textit{-0.29}} & \textbf{\textit{-1.32}} & \textbf{\textit{-0.33}} & \textit{-4.06} & \textbf{\textit{-0.78}} & \textit{-2.32} & \textbf{\textit{-0.38}} \\
\midrule
\multicolumn{13}{l}{\textit{\textbf{Backbone: Qwen3-VL}}} \\
Vanilla (Full Token Bound) & 4B & 3193 & 72.04 & \textbf{70.52} & \textbf{54.83} & 93.26 & \textbf{68.55} & 78.61 & \textbf{57.24} & 68.57 & \textbf{60.42} & \textbf{69.34} \\
Fourier-Qwen-3 (Spatial)& 4B & 1328 & 70.81 & 69.04 & 52.75 & 92.23 & 61.02 & 76.65 & 54.21 & 66.08 & 60.67 & 66.92 \\
\textit{$\Delta$ vs. Vanilla} & & \textit{-58.4\%} & \textit{-1.23} & \textit{-1.48} & \textit{-2.08} & \textit{-1.03} & \textit{-7.53} & \textit{-1.96} & \textit{-3.03} & \textit{-2.49} & \textit{+0.25} & \textit{-2.42} \\
\rowcolor{blue!5} \textbf{Fre-Res-Compact (Ours)} & 4B & \textbf{1280} & \textbf{73.28} & 70.43 & 54.55 & \textbf{93.48} & 68.32 & \textbf{78.65} & 53.47 & \textbf{68.69} & 60.21 & 69.06 \\
\rowcolor{blue!5} \textit{$\Delta$ vs. Vanilla} & & \textbf{\textit{-59.9\%}} & \textbf{\textit{+1.24}} & \textbf{\textit{-0.09}} & \textbf{\textit{-0.28}} & \textbf{\textit{+0.22}} & \textbf{\textit{-0.23}} & \textbf{\textit{+0.04}} & \textit{-3.77} & \textbf{\textit{+0.12}} & \textit{-0.21} & \textbf{\textit{-0.28}} \\
\bottomrule
\end{tabular}
}
\end{table}

The controlled results show a consistent advantage for Fre-Res under stricter token budgets. 
On Qwen2-VL, Fre-Res-Compact uses slightly fewer visual tokens than Fourier-Qwen-2, 1280 versus 1328, while improving the average accuracy from 59.86 to 61.05.
It also stays close to the full-token vanilla model, which reaches 61.43 with 3193 visual tokens. 
The same pattern holds on Qwen3-VL: Fre-Res-Compact obtains 69.06 average accuracy with 1280 visual tokens, nearly matching the full-token vanilla model at 69.34 and clearly outperforming Fourier-Qwen-3 at 66.92.

The category-level results further reveal the difference between spatial and temporal compression. 
Spatial frequency truncation can reduce the token count effectively, but it tends to lose more information in categories that depend on spatial layout or object multiplicity. 
For example, on Qwen2-VL, Fourier-Qwen-2 drops by 4.81 points on Position and 5.32 points on Count, while Fre-Res-Compact reduces these drops to 0.71 and 1.32 points. 
On Qwen3-VL, the Count drop is reduced from 7.53 points with Fourier compression to only 0.23 points with Fre-Res. 
This is consistent with our design: Fre-Res does not directly truncate spatial frequency components within each frame, but preserves spatial anchors and compresses the temporal residuals between them.

Fre-Res-Compact is also competitive on action-centric questions. 
On Qwen2-VL and Qwen3-VL, it slightly exceeds the full-token vanilla models on Action, with gains of 0.35 and 1.24 points. 
We do not interpret this as compression universally improving the backbone. 
Rather, it indicates that explicitly modeling inter-frame residuals can make motion-related evidence easier for the LLM to access, while reducing the amount of redundant static visual information.

The main remaining weakness is Pose. 
Although Fre-Res-Compact better preserves Position and Count than Fourier-based compression on the Qwen-family backbones, its Pose accuracy still drops compared with the full-token vanilla models. 
This is expected, since pose reasoning often relies on subtle local spatial configurations and high-frequency visual details. 
These details are not always fully retained by sparse spatial anchors and temporal residual compression alone. 
Improving anchor selection for pose-sensitive scenarios is therefore an important direction for future work.

\subsection{Temporal Reasoning and Long-Video Understanding (RQ2)}
\label{subsec:long_video}

Beyond short-video spatial reasoning, we further evaluate whether Fre-Res can preserve temporal evidence under longer video contexts. 
This setting is particularly important for our method, since the temporal residual branch is designed to retain inter-frame dynamics while reducing redundant visual tokens. 
We evaluate on NExT-QA, LongVideoBench, and EgoSchema, which cover complementary aspects of temporal reasoning, long-context video understanding, and egocentric event comprehension. 
Unlike MVBench in Table~\ref{tab:mvbench}, these benchmarks are reported using standard accuracy.

\begin{table*}[t]
\centering
\caption{\textbf{Accuracy on temporal reasoning and long-video benchmarks.}
We evaluate Fre-Res on NExT-QA, LongVideoBench, and EgoSchema to measure temporal reasoning and long-context video understanding.
Fre-Res in this table uses a longer-context budgeted instantiation of the same dual-track framework, with a larger visual-token budget than the compact MVBench setting to better preserve event-level temporal evidence.
All scores are reported as accuracy (\%).
Visual Tokens denotes the average number of visual embeddings injected into the LLM after compression/projection, excluding text and special prompt tokens.
Compr. is computed relative to the corresponding full-token vanilla model.
$\Delta$ is reported in percentage points relative to the vanilla model.}
\label{tab:long_video}
\resizebox{\textwidth}{!}{
\begin{tabular}{l c c | c c c}
\toprule
\textbf{Method} & \textbf{Visual Tokens} & \textbf{Compr.} 
& \textbf{NExT-QA Acc.} 
& \textbf{LongVideoBench Acc.} 
& \textbf{EgoSchema Acc.} \\
\midrule
\multicolumn{6}{l}{\textit{\textbf{Backbone: Qwen2-VL-7B}}} \\
Vanilla (Full Token Bound) & $\sim$14,000 & 1.0$\times$ & \textbf{71.24} & \textbf{55.61} & \textbf{66.73} \\
DyCoke (Attention-based Drop) & $\sim$6,600 & 2.1$\times$ & 65.42 & 46.93 & 55.86 \\
CDPruner (Similarity-based Merge) & $\sim$6,600 & 2.1$\times$ & 66.18 & 48.37 & 57.31 \\
Fourier-VLM (Spatial Compression) & $\sim$6,600 & 2.1$\times$ & 67.05 & 49.18 & 58.24 \\
\rowcolor{blue!5} \textbf{Fre-Res (Ours)} & \textbf{$\sim$6,600} & \textbf{2.1$\times$} & 70.08 & 53.52 & 63.58 \\
\rowcolor{blue!5} \textit{$\Delta$ vs. Vanilla} & & & \textbf{\textit{-1.16}} & \textbf{\textit{-2.09}} & \textbf{\textit{-3.15}} \\
\midrule
\multicolumn{6}{l}{\textit{\textbf{Backbone: Qwen3-VL-8B}}} \\
Vanilla (Full Token Bound) & $\sim$16,000 & 1.0$\times$ & \textbf{79.62} & \textbf{63.74} & \textbf{67.38} \\
DyCoke (Attention-based Drop) & $\sim$7,600 & 2.1$\times$ & 74.35 & 55.28 & 56.84 \\
CDPruner (Similarity-based Merge) & $\sim$7,600 & 2.1$\times$ & 75.06 & 56.71 & 58.26 \\
Fourier-VLM (Spatial Compression) & $\sim$7,600 & 2.1$\times$ & 75.91 & 57.64 & 59.08 \\
\rowcolor{blue!5} \textbf{Fre-Res (Ours)} & \textbf{$\sim$7,600} & \textbf{2.1$\times$} & 78.74 & 61.96 & 64.22 \\
\rowcolor{blue!5} \textit{$\Delta$ vs. Vanilla} & & & \textbf{\textit{-0.88}} & \textbf{\textit{-1.78}} & \textbf{\textit{-3.16}} \\
\bottomrule
\end{tabular}
}
\end{table*}

As shown in Table~\ref{tab:long_video}, Fre-Res consistently achieves the best performance among the compressed methods under the same compression ratio. 
On Qwen2-VL-7B, all compressed methods use approximately 6.6K visual tokens, corresponding to a 2.1$\times$ reduction from the full-token vanilla model. 
Under this matched budget, Fre-Res obtains 70.08 on NExT-QA, 53.52 on LongVideoBench, and 63.58 on EgoSchema, outperforming DyCoke, CDPruner, and Fourier-VLM across all three benchmarks. 
The gap is especially clear on LongVideoBench and EgoSchema, where Fre-Res improves over Fourier-VLM by 4.34 and 5.34 points, respectively.

The same trend holds on Qwen3-VL-8B. 
With approximately 7.6K visual tokens, Fre-Res reaches 78.74 on NExT-QA, 61.96 on LongVideoBench, and 64.22 on EgoSchema. 
Compared with Fourier-VLM, this corresponds to gains of 2.83, 4.32, and 5.14 points. 
At the same time, Fre-Res remains close to the full-token vanilla model, with drops of only 0.88 points on NExT-QA and 1.78 points on LongVideoBench. 
The larger drop on EgoSchema indicates that egocentric long-horizon reasoning remains more sensitive to compression, but Fre-Res still preserves substantially more accuracy than the other compressed baselines.

\begin{figure}[t]
    \centering
    \includegraphics[width=0.8\textwidth]{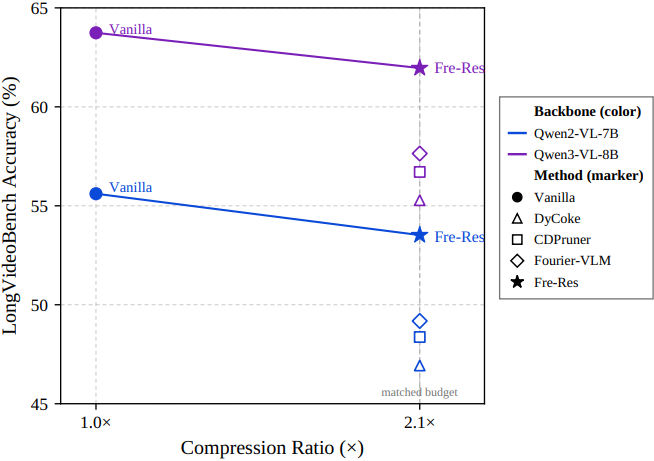}
   \caption{\textbf{Accuracy--efficiency trade-off on LongVideoBench.}
Fre-Res achieves a favorable trade-off compared with attention-based dropping, similarity-based merging, and spatial frequency compression under the same matched compression ratio.
Each color denotes a backbone, and each marker denotes a compression method.
While the full-token vanilla model obtains the highest accuracy, Fre-Res retains most of its performance with approximately $2.1\times$ fewer visual tokens.}
\label{fig:pareto}
\end{figure}

These results suggest that temporal compression should not be treated as a generic token-pruning problem. 
Attention-based dropping and similarity-based merging can reduce sequence length, but they do not explicitly model the temporal residual structure of video. 
Fourier-VLM introduces frequency-domain compression, yet its compression is primarily spatial and therefore does not directly capture inter-frame evolution. 
In contrast, Fre-Res preserves high-fidelity spatial anchors while representing temporal changes through compact residual-frequency tokens. 
This design better matches the structure of long-video reasoning, where the model must retain both event-level continuity and sufficient spatial evidence. We further visualize this accuracy--efficiency comparison in Figure~\ref{fig:pareto}.
Under the matched compressed budget, Fre-Res forms the strongest compressed point for both Qwen2-VL-7B and Qwen3-VL-8B, retaining most of the full-token performance while reducing the visual-token budget by approximately $2.1\times$.

\subsection{Hardware Profiling and Practical Scalability (RQ3)}
\label{subsec:efficiency}

\begin{table}[t]
\centering
\caption{\textbf{Empirical hardware profiling on a single H100 GPU.}
We measure real-world inference overhead on a 1-minute dense video with 30 FPS, corresponding to 1800 frames, using Qwen3-VL-8B with FlashAttention-2.
Fre-Res in this table refers to the long-video instantiation of the same budget-adaptive framework, where the temporal aggregation range is enlarged to keep the total sequence length practical for dense high-frame-rate inputs.
Total Context Tokens includes visual embeddings and approximately 22 text/system tokens.
Peak VRAM is measured with \texttt{torch.cuda} and includes BF16 model weights, KV cache, and runtime activations.
TTFT denotes Time-To-First-Token latency for the prefill stage.
Fourier applies frame-wise spatial compression and therefore still scales with the number of frames, while Fre-Res reduces the input to sparse spatial anchors and compressed temporal residuals. The reported TTFT is the LLM-side prefill latency after visual embeddings are constructed; model-side preprocessing overhead is discussed below.}
\label{tab:hardware}
\resizebox{\columnwidth}{!}{
\begin{tabular}{l c c c}
\toprule
\textbf{Mechanism / Strategy} & \textbf{Total Context Tokens} & \textbf{Peak VRAM (GB)} & \textbf{TTFT (s)} \\
\midrule
Vanilla (Dense 30 FPS) & 1,036,822 & \textbf{OOM} & \textbf{OOM} \\
Uniform Sampling (1 FPS) & 34,582 & 19.42 & 1.15 \\
Fourier (Frame-wise Spatial 8$\times$) & 129,622 & 41.68 & 7.84 \\
\rowcolor{blue!5} \textbf{Fre-Res (Ours)} & \textbf{46,102}$^\dagger$ & \textbf{22.35} & \textbf{1.68} \\
\bottomrule
\end{tabular}
}
\vspace{0.5mm}
\footnotesize{
$^\dagger$ Fre-Res uses sparse raw anchors plus compressed temporal residual tokens; the total includes approximately 22 text/system tokens.
}
\end{table}

We further evaluate whether the token reduction of Fre-Res translates into practical system-level savings. 
To this end, we profile inference on a single H100 GPU using Qwen3-VL-8B with FlashAttention-2. 
The input is a 1-minute dense video sampled at 30 FPS, resulting in 1800 frames. 
This setting is intentionally challenging: directly feeding all frames produces more than one million context tokens and exceeds the memory limit even on an H100.

As shown in Table~\ref{tab:hardware}, the dense vanilla setting leads to out-of-memory failure, confirming that naive long-video ingestion is impractical at high frame rates. 
Frame-wise Fourier compression reduces the spatial token count per frame, but it still processes every frame independently. 
As a result, its total context length remains high at 129,622 tokens, requiring 41.68 GB peak VRAM and 7.84 seconds of TTFT.

We additionally measure the model-side preprocessing overhead under the same H100 setting, excluding video decoding and disk I/O.
Since both Fourier and Fre-Res encode all 1800 frames through the visual encoder, their visual-encoding overhead is comparable.
Fre-Res adds only lightweight operations after visual encoding, including residual computation, temporal DCT, and the Spatial-Guided Absorber.
In our implementation, these additional operations add about 0.04\,s beyond visual encoding.
This overhead is small compared with the LLM-side prefill latency: Fre-Res reduces TTFT from 7.84\,s to 1.68\,s compared with Fourier, corresponding to a 6.16\,s reduction at the LLM stage.
Therefore, the primary efficiency gain of Fre-Res comes from reducing the number of visual tokens injected into the LLM, rather than from preprocessing savings.

Fre-Res substantially lowers this overhead by replacing dense per-frame tokenization with sparse spatial anchors and compressed temporal residuals. 
It reduces the total context length to 46,102 tokens, which is about 2.8$\times$ fewer tokens than Fourier compression and about 22.5$\times$ fewer tokens than the dense vanilla input. 
This reduction directly translates into lower runtime cost: peak VRAM decreases from 41.68 GB to 22.35 GB compared with Fourier, and TTFT drops from 7.84 seconds to 1.68 seconds.

Uniform 1 FPS sampling has the lowest hardware cost in this table, but it achieves this by discarding 29 out of every 30 frames. 
This makes it efficient but vulnerable to missing short-lived events and dense temporal transitions. 
In contrast, Fre-Res is designed for dense-video understanding: it keeps sparse high-fidelity spatial anchors while encoding the remaining temporal evolution through residual-frequency tokens. 
Therefore, the main advantage of Fre-Res is not simply minimizing token count, but providing a better practical trade-off between temporal coverage, memory usage, and prefill latency.
\subsection{Component Analysis via Extreme Ablation (RQ4)}
\label{subsec:ablation}

While Sections~\ref{subsec:rq1_mvbench} and~\ref{subsec:long_video} evaluate the end-to-end accuracy--efficiency trade-off of Fre-Res, this section deconstructs the internal roles of its two complementary tracks.
Specifically, we ask three questions:
\textit{(i)} whether the temporal-frequency residuals alone carry usable causal evidence;
\textit{(ii)} whether spatial anchors remain useful when the temporal branch is already strong; and
\textit{(iii)} whether the proposed Spatial-Guided Absorber is effective for aligning frequency-domain dynamics with spatial visual tokens.
We conduct the main ablation on NExT-QA, a benchmark that emphasizes temporal transitions, causal relations, and event-level dynamics.
To further verify the complementary roles of the two tracks across task types, we provide an additional cross-benchmark ablation on representative MVBench categories in Appendix~\ref{app:cross_ablation}.

\begin{table}[t]
\centering
\caption{\textbf{Extreme ablation on causal temporal reasoning.}
We evaluate branch-level and fusion-level variants on NExT-QA using Qwen2-VL-7B.
The full-token 16-frame baseline obtains 71.24\% accuracy.
The temporal-frequency track performs strongly on this temporal-heavy benchmark, suggesting that latent 1D-DCT residuals preserve compact action-transition evidence.
Combining it with spatial anchors gives the best performance, while the Spatial-Guided Absorber substantially outperforms naive concatenation and iDCT-based reconstruction.}
\label{tab:ablation}
\resizebox{\columnwidth}{!}{
\begin{tabular}{l c c}
\toprule
\textbf{Configuration} & \textbf{Acc. (\%)} & \textbf{$\Delta$} \\
\midrule
\multicolumn{3}{l}{\textit{Phase 1: Branch-level decomposition}} \\
(A) Spatial Anchor Track Only (8 raw anchors) 
& 65.59 & -- \\
(B) Temporal-Frequency Track Only (1D-DCT residuals) 
& 68.47 & \textit{+2.88 vs. A} \\
\rowcolor{blue!5} 
\textbf{(C) Dual-Track Fre-Res} 
& \textbf{70.08} & \textbf{\textit{+4.49 vs. A / +1.61 vs. B}} \\
\midrule
\multicolumn{3}{l}{\textit{Phase 2: Fusion-level decomposition}} \\
(D) Dual-Track + Linear Concatenation 
& 66.42 & -- \\
(E) Dual-Track + iDCT Reconstruction 
& 67.85 & \textit{+1.43 vs. D} \\
\rowcolor{blue!5} 
\textbf{(F) Dual-Track + Spatial-Guided Absorber (Ours)} 
& \textbf{70.08} & \textbf{\textit{+3.66 vs. D / +2.23 vs. E}} \\
\bottomrule
\end{tabular}
}
\end{table}

\begin{figure}[t]
    \centering
    \includegraphics[width=\textwidth]{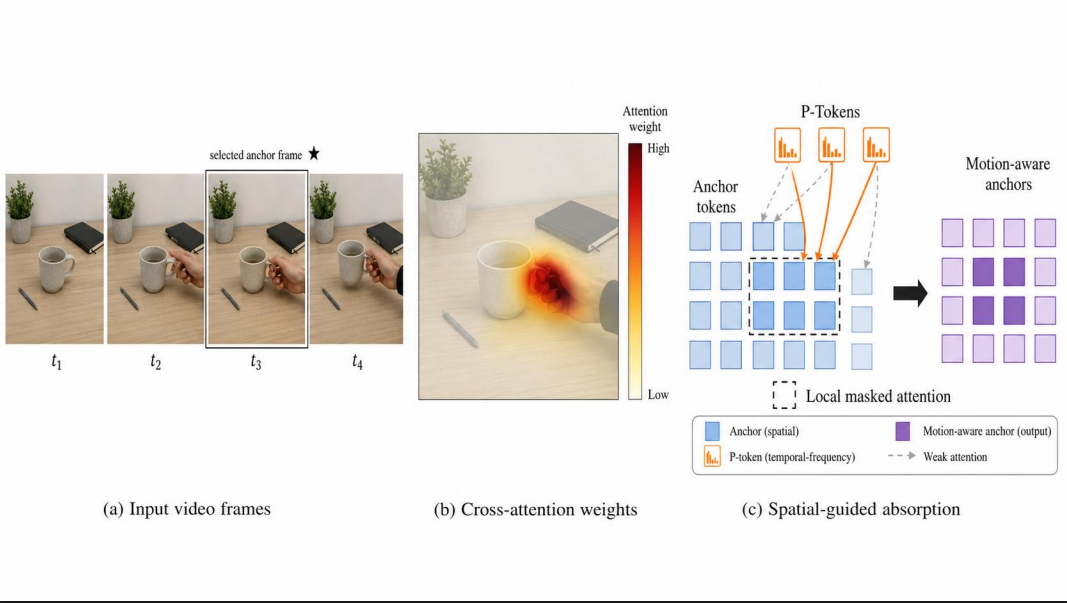}
    \caption{\textbf{Qualitative visualization and schematic illustration of the Spatial-Guided Absorber.}
\textbf{(a)} Input video frames, where the selected anchor frame is highlighted.
\textbf{(b)} Cross-attention weights visualized on the selected anchor frame.
Dynamic regions around the hand and cup receive stronger attention, while static background regions receive weaker attention.
\textbf{(c)} Schematic illustration of spatial-guided absorption.
Anchor tokens locally attend to nearby temporal-frequency P-Tokens, producing motion-aware anchors in dynamic regions while leaving mostly unchanged background anchors with weak responses.}
\label{fig:attention}

\end{figure}

Figure~\ref{fig:attention} provides a qualitative view of how the Spatial-Guided Absorber behaves.
In the selected anchor frame, the cross-attention weights concentrate around the dynamic hand--cup interaction region, while the static background receives weaker responses.
This pattern is consistent with the intended role of the absorber: temporal-frequency P-Tokens are not globally injected into all spatial tokens, but are routed toward local regions where residual dynamics are more relevant.
The schematic in Figure~\ref{fig:attention}(c) further illustrates this local absorption process, where anchor tokens in dynamic regions become motion-aware while mostly unchanged background anchors remain weakly affected.

\textbf{Track-level decomposition.}
Table~\ref{tab:ablation} reveals a clear benchmark-dependent behavior.
On NExT-QA, the Temporal-Frequency Track alone reaches 68.47\%, outperforming the Spatial Anchor Track by 2.88 points.
This result is consistent with the nature of NExT-QA: many questions require identifying causal transitions, action evolution, and event consequences, for which compact inter-frame residual dynamics are highly informative.
Therefore, the strong performance of the temporal-only branch should not be interpreted as an anomaly.
Instead, it suggests that latent 1D-DCT residuals carry discriminative information beyond 
compressed noise, as evidenced by their strong performance on temporal reasoning tasks.

However, this strong temporal-only result is benchmark-dependent and does not eliminate the need for spatial anchors.
The additional cross-benchmark ablation in Appendix~\ref{app:cross_ablation} verifies this contrast more directly.
The Temporal-Frequency Track remains strong on NExT-QA but degrades sharply on spatially sensitive MVBench categories such as Count and Pose, whereas the Spatial Anchor Track provides much stronger spatial reasoning performance.
This contrast reveals a clear division of labor between the two tracks.
The Temporal-Frequency Track is effective at preserving causal dynamics and inter-frame transitions, whereas the Spatial Anchor Track provides high-fidelity object-level and layout-level evidence.
Thus, the Raw Anchor Branch is not redundant; it compensates for the spatial blindness of pure residual-frequency compression.
Combining both tracks yields the best NExT-QA accuracy of 70.08\%, indicating that spatial evidence remains complementary even on a temporal-heavy benchmark.

\textbf{Fusion-level decomposition.}
A naive linear concatenation of spatial anchors and temporal-frequency tokens obtains only 66.42\% on NExT-QA.
The same trend is also observed in the cross-benchmark ablation in Appendix~\ref{app:cross_ablation}, where the Spatial-Guided Absorber improves over linear concatenation on MVBench.
This result suggests that directly appending frequency-domain residual tokens to the visual sequence is insufficient for a frozen LLM, since these tokens lie on a different representational manifold from native visual embeddings.
In other words, the main difficulty is not only whether temporal residuals are informative, but also whether their frequency-domain semantics can be presented to the LLM in an interpretable form.

Reconstructing the residuals back to the spatial domain through iDCT improves the accuracy to 67.85\%, showing that inverse transformation can partially reduce the semantic gap.
However, iDCT reconstruction still remains clearly below the proposed absorber.
This indicates that mapping residuals back to dense spatial grids may reintroduce redundant visual tokens and blur the compact temporal-frequency evidence.
In contrast, the Spatial-Guided Absorber reaches 70.08\%, improving over naive concatenation by 3.66 points and over iDCT reconstruction by 2.23 points.
By using local masked cross-attention, the absorber injects temporal-frequency dynamics into spatially corresponding anchor tokens before language decoding.
This enables the LLM to consume motion-aware spatial anchors rather than raw frequency tokens, thereby providing an effective latent-space realization of motion-aware fusion.

Overall, the ablation demonstrates that Fre-Res does not simply reduce visual tokens with a single pruning rule.
Instead, it separates video evidence into two complementary forms: spatial anchors for fine-grained visual fidelity and temporal-frequency residuals for causal dynamics.
The final performance gain arises from both preserving these complementary signals and aligning them through the Spatial-Guided Absorber.

\section{Conclusion}
\label{sec:conclusion}

In this paper, we introduced \textbf{Fre-Res}, a dual-track video-token compression framework for efficient video MLLMs.
Fre-Res separates video evidence into high-fidelity spatial anchors and compact temporal-frequency residuals, preserving fine-grained visual details and dense inter-frame dynamics under a reduced token budget.
By applying temporal 1D-DCT to latent residual trajectories, Fre-Res exploits the low-frequency concentration of video dynamics and provides a structured alternative to homogeneous token pruning.

Our experiments show that temporal-frequency residuals are especially effective for causal video reasoning, while raw spatial anchors remain essential for fine-grained spatial tasks such as position, counting, object, attribute, and pose understanding.
This confirms the complementary roles of the two tracks.
Moreover, the proposed Spatial-Guided Absorber effectively bridges the semantic gap between frequency-domain residuals and native visual tokens, outperforming naive concatenation and iDCT-based reconstruction.

Across short-video and long-video benchmarks, Fre-Res achieves a favorable accuracy--efficiency trade-off, substantially reducing visual-token length while maintaining competitive reasoning performance.
These results suggest that scalable video MLLMs should compress video according to its structured spatiotemporal redundancy rather than relying on unstructured token dropping.
Future work will explore adaptive anchor allocation, query-aware frequency selection, and hierarchical compression for hour-scale video understanding.

\newpage

{
\small
\bibliographystyle{plain}
\bibliography{bible}

}


\newpage
\appendix

\section{Appendix}

\section{Limitations}
\label{app:limitations}

Fre-Res provides a structured way to reduce visual-token length by separating spatial anchors from temporal-frequency residuals, but several limitations remain.

\textbf{Pose-sensitive and fine-grained local reasoning.}
Although Fre-Res preserves local spatial coverage through raw anchors, our experiments show that pose-related questions remain more sensitive to compression.
This is expected because pose reasoning often depends on subtle local configurations, small articulated parts, and high-frequency visual details.
A fixed anchor allocation may not always retain the most informative regions for these tasks.
Future work may incorporate adaptive anchor allocation or query-aware token retention to better handle pose-sensitive scenarios.

\textbf{Dependence on temporal smoothness.}
The temporal-frequency branch assumes that short-range latent residual trajectories often exhibit low-frequency concentration.
This assumption is generally valid for many natural videos, but may be less effective for abrupt scene cuts, fast camera motion, flashing patterns, or highly discontinuous edits.
In such cases, more anchors or shorter GoP intervals may be required to avoid losing high-frequency temporal evidence.

\textbf{Dataset-dependent branch behavior.}
Our ablations show that the temporal-frequency track performs strongly on NExT-QA, where causal and temporal transitions are central.
However, the same temporal-only branch performs poorly on spatially sensitive MVBench categories.
This indicates that the two tracks are complementary rather than interchangeable.
The optimal allocation between spatial anchors and temporal residual tokens may therefore depend on the benchmark, task type, and input video distribution.

\textbf{Budget-adaptive configuration.}
Fre-Res is not a single fixed-token configuration.
Different experiments instantiate the same dual-track design under different visual-token budgets.
While this allows fair comparisons under diverse settings, it also introduces additional configuration choices, including anchor count, spatial token-retention ratio, GoP size, and temporal aggregation range.
We report the actual visual-token count in each experiment, but a more systematic study of budget allocation remains future work.

\textbf{Long-horizon video understanding.}
For dense long videos, Fre-Res substantially reduces token length, but hour-scale video understanding may still require hierarchical temporal summarization beyond GoP-level residual compression.
Combining Fre-Res with memory mechanisms, event-level indexing, or hierarchical video summarization is a promising direction.

\section{Ethics Statement}
\label{app:ethics}

Fre-Res is a video-token compression framework for improving the efficiency of video MLLMs.
It does not introduce new data collection procedures and does not require annotating personally identifiable information beyond what is already present in the evaluated public benchmarks.
However, because Fre-Res operates on video inputs, it inherits the ethical concerns of video understanding systems.

\textbf{Privacy and surveillance.}
More efficient video understanding may reduce the computational barrier for processing long videos.
While this can enable beneficial applications such as assistive video search, educational video analysis, and efficient robotics perception, it may also be misused for large-scale surveillance or privacy-invasive monitoring.
Deployment should therefore follow appropriate privacy policies, consent requirements, and data protection regulations.

\textbf{Bias and representational harms.}
Fre-Res uses frozen pretrained video MLLM backbones.
As a result, it may inherit biases already present in the underlying visual encoders, language models, and training data.
Compression may also amplify errors if important evidence is removed or underrepresented.
For sensitive applications involving people, identity, behavior, or safety-critical decisions, additional auditing and human oversight are necessary.

\textbf{Reliability under compression.}
Token compression can change model predictions by removing or transforming visual evidence.
Although Fre-Res is designed to preserve spatial anchors and temporal residual dynamics, it may still fail on videos with abrupt motion, small objects, rare events, or ambiguous causal relations.
Users should avoid treating compressed-model outputs as guaranteed faithful summaries of the original video.

\textbf{Responsible release.}
If code and checkpoints are released, we recommend clearly documenting supported use cases, known failure modes, expected compression settings, and benchmark-specific configurations.
We also encourage reporting both accuracy and token-efficiency metrics to avoid overclaiming model capability under aggressive compression.

\section{Baseline Fairness and Budget Matching}
\label{app:baseline_fairness}

For all controlled comparisons in Tables~\ref{tab:mvbench} and~\ref{tab:long_video}, we use the same backbone, prompt format, decoding configuration, and answer extraction rule within each benchmark.
For each compression baseline, we follow the officially reported hyperparameters whenever available and adjust only the primary budget-control parameter to match the target visual-token count.
Specifically, we adjust the attention retention threshold for DyCoke, the similarity threshold for CDPruner, and the frequency truncation level for Fourier-VLM.

For the temporal and long-video benchmarks in Table~\ref{tab:long_video}, the target visual-token counts are approximately 6,600 for Qwen2-VL-7B and 7,600 for Qwen3-VL-8B.
For MVBench in Table~\ref{tab:mvbench}, Fre-Res-Compact is constrained to use no more visual tokens than the corresponding Fourier-based spatial compression baseline.
No task-specific re-training is applied to DyCoke, CDPruner, Fourier-VLM, or other compression baselines.

To verify that the reported operating points are not artifacts of a single budget choice, we additionally evaluate each baseline under nearby token budgets by varying its primary budget-control parameter.
In our sweep, we consider approximately $\pm20\%$ visual-token budgets around the reported operating point.
Across these nearby settings, we do not observe substantial accuracy improvements over the reported matched-budget results under the same backbone.

\section{Training Details}
\label{app:training_details}

\subsection{Trainable Components}

In all Fre-Res variants, the visual encoder and LLM backbone are frozen unless otherwise specified.
Only lightweight compression and fusion modules are optimized.
The trainable components include:

\begin{itemize}[leftmargin=*]
    \item \textbf{RawAdapter:} a lightweight adapter applied to raw-anchor visual tokens to align their distribution with the multimodal fusion space.
    \item \textbf{Fre-Res token adapters:} projection layers for dynamic anchors, static tokens, temporal-frequency P-Tokens, and frequency-summary tokens.
    \item \textbf{Spatial-Guided Absorber:} local masked cross-attention layers that align temporal-frequency residuals with spatial anchor tokens.
    \item \textbf{Type embeddings:} learnable embeddings indicating whether each token comes from the raw spatial branch or the Fre-Res temporal branch.
    \item \textbf{Branch gates:} learnable scalar gates used to balance the magnitude of spatial-anchor tokens and temporal-frequency tokens.
\end{itemize}

\subsection{Training Objective}

The primary training objective is the standard autoregressive language modeling loss over the target answer tokens:
\begin{equation}
    \mathcal{L}_{\mathrm{task}}
    =
    - \sum_{t=1}^{L}
    \log p_{\theta}(y_t \mid y_{<t}, H_{\mathrm{vis}}, q),
\end{equation}
where $H_{\mathrm{vis}}$ denotes the compressed visual-token sequence, $q$ denotes the text question or instruction, and $y_t$ is the target answer token.

When knowledge distillation is used, we additionally minimize the divergence between the option logits of the compressed student model and the full-token teacher model:
\begin{equation}
    \mathcal{L}_{\mathrm{KD}}
    =
    \mathrm{KL}
    \left(
    \mathrm{softmax}(z_{\mathrm{teacher}} / \tau)
    \,\|\, 
    \mathrm{softmax}(z_{\mathrm{student}} / \tau)
    \right),
\end{equation}
where $\tau$ is the distillation temperature.

The total objective is:
\begin{equation}
    \mathcal{L}
    =
    \mathcal{L}_{\mathrm{task}}
    +
    \lambda_{\mathrm{KD}} \mathcal{L}_{\mathrm{KD}}
    +
    \lambda_{\mathrm{gate}} \mathcal{L}_{\mathrm{gate}},
\end{equation}
where $\mathcal{L}_{\mathrm{gate}}$ is an optional regularization term for stabilizing branch gates or frequency selection.
Unless otherwise specified, MVBench is used only for evaluation and is not included in training.

\section{Budget Allocation Details}
\label{app:budget}

Fre-Res is a budget-adaptive framework.
Instead of using a fixed number of output visual tokens for all inputs, it instantiates the same dual-track design under different target budgets by adjusting the number of spatial anchors, the number of raw tokens retained per anchor, and the temporal aggregation range.
This section details the budget allocation rule used in our experiments.

\subsection{Hierarchical Budget Allocation}

Given a target visual-token budget $B$, Fre-Res allocates tokens in a hierarchical manner.
We first reserve a budget for the raw spatial-anchor branch:
\begin{equation}
    B_{\mathrm{spatial}} = M \cdot K_{\mathrm{raw}},
\end{equation}
where $M$ is the number of selected anchor frames and $K_{\mathrm{raw}}$ is the number of retained raw visual tokens per anchor frame.
For the illustrative 16-frame configuration in Figure~\ref{fig:architecture}, we use $M=8$ anchors and retain $K_{\mathrm{raw}}=512$ tokens per anchor, corresponding to $4096$ spatial-anchor tokens.
For compact budget-matched settings such as MVBench, $M$ and $K_{\mathrm{raw}}$ are reduced so that the total visual-token count does not exceed the target budget of the corresponding compression baseline.

After reserving spatial-anchor tokens, the remaining budget is assigned to the temporal Fre-Res branch:
\begin{equation}
    B_{\mathrm{freq}}
    =
    B
    -
    B_{\mathrm{spatial}}
    -
    B_{\mathrm{summary}}
    -
    B_{\mathrm{static}},
\end{equation}
where $B_{\mathrm{summary}}$ denotes the budget for global frequency-summary tokens and $B_{\mathrm{static}}$ denotes the budget for static background tokens.

Within $B_{\mathrm{freq}}$, the number of retained temporal DCT coefficients is selected according to the available budget:
\begin{equation}
    K
    =
    \min
    \left(
    K_{\max},
    L_{\mathrm{group}},
    \left\lfloor
    \frac{B_{\mathrm{freq}}}
    {N_{\mathrm{group}} \cdot N_{\mathrm{pool}}}
    \right\rfloor
    \right),
\end{equation}
where $N_{\mathrm{group}}$ is the number of temporal groups, $N_{\mathrm{pool}}$ is the number of pooled spatial positions used by the temporal-frequency branch, and $L_{\mathrm{group}}$ is the effective temporal length of each group.
The cap by $L_{\mathrm{group}}$ ensures that the number of retained DCT coefficients does not exceed the available temporal resolution.
The cap by $K_{\max}$ prevents the temporal branch from allocating excessive coefficients when the budget is loose.

\subsection{Compact 16-frame Instantiation}

In the compact 16-frame instantiation, the Fre-Res temporal branch produces approximately 317 tokens.
After temporal grouping and spatial pooling, each temporal group produces a small set of candidate P-Tokens from the retained low-frequency residual coefficients.
In our compact setting, we use 8 temporal groups, retain $K=3$ low-frequency coefficients, and pool residuals onto a $4\times4$ spatial grid, producing up to
\begin{equation}
    8 \times 3 \times 16 = 384
\end{equation}
candidate P-Tokens before energy-based filtering.
After removing near-zero-energy positions, the number of dynamic P-Tokens is reduced to approximately 285 on average.
Together with 8 global frequency-summary tokens ($P_{\mathrm{summary}}$, one per temporal group) and approximately 24 static background tokens ($S$), the Fre-Res branch yields about
\begin{equation}
    285 + 8 + 24 \approx 317
\end{equation}
tokens.

\subsection{Budgeted Experimental Instantiations}

Different experiments use different instantiations of the same allocation rule.

For MVBench, we use a compact budget-matched setting.
The anchor count and raw-token retention per anchor are reduced so that Fre-Res-Compact uses no more visual tokens than the corresponding spatial-compression baseline.
This allows a conservative comparison under a stricter token budget.

For temporal and long-video reasoning benchmarks, we use a larger visual-token budget to preserve event-level temporal evidence.
The temporal aggregation range is increased, and the Fre-Res branch allocates more residual-frequency evidence while maintaining sparse spatial anchors.

For dense long-video hardware profiling, the temporal aggregation range is further enlarged and the anchor density is adjusted to keep the total sequence length practical for high-frame-rate inputs.
This allows Fre-Res to reduce million-token dense video inputs to a bounded long-context sequence while retaining both sparse spatial anchors and compact temporal-frequency evidence.

\subsection{Implementation Notes}

For multiple-choice video QA, we include all answer options in the text prompt during training and evaluation.
This avoids distribution mismatch between training and evaluation and ensures that the model predicts among the provided choices.
During generation with custom visual embeddings, care is taken to decode only the newly generated answer tokens.
All reported visual-token counts exclude text tokens and special prompt tokens unless otherwise specified.

\subsection{Hyperparameter Sensitivity}
\label{app:sensitivity}

We further study the sensitivity of Fre-Res to several key hyperparameters.
Table~\ref{tab:sensitivity} reports NExT-QA accuracy under different temporal grouping scales $G$, retained DCT coefficient budgets $K$, and spatial anchor counts $M$ using Qwen2-VL-7B.
Here, $G$ denotes the temporal grouping scale used for residual construction, while the DCT is applied over the effective residual trajectory within each group.
The reported token count refers only to the compact Fre-Res temporal branch, excluding raw spatial-anchor tokens and text tokens.

The default configuration $(G{=}2,K{=}3,M{=}8)$ achieves the best accuracy in this sweep.
Nearby configurations remain competitive, suggesting that Fre-Res is not overly sensitive to a single hyperparameter choice.
Increasing the coefficient budget to $K{=}5$ or the anchor count to $M{=}10$ does not consistently improve performance, indicating that simply allocating more frequency coefficients or anchors is not always beneficial.

\begin{table}[t]
\centering
\caption{\textbf{Sensitivity of Fre-Res to key hyperparameters on NExT-QA.}
All variants use Qwen2-VL-7B.
We vary the temporal grouping scale ($G$), retained DCT coefficients ($K$), and spatial anchor count ($M$).
Fre-Res Tokens denote compact temporal-branch tokens after energy-based filtering, excluding raw spatial anchors and text tokens.}
\label{tab:sensitivity}
\small
\begin{tabular}{ccc|cc}
\toprule
$G$ & $K$ & $M$ & \textbf{NExT-QA Acc. (\%)} & \textbf{Fre-Res Tokens$^\dagger$} \\
\midrule
2 & 1 & 8 & 68.34 & 190 \\
\rowcolor{blue!5} 
2 & 3 & 8 & \textbf{70.08} & 317 \\
4 & 2 & 8 & 67.55 & 184 \\
2 & 5 & 8 & 69.41 & 491 \\
2 & 3 & 6 & 68.62 & 238 \\
2 & 3 & 10 & 69.85 & 396 \\
\bottomrule
\end{tabular}\\
\vspace{1mm}
\raggedright
\footnotesize{$^\dagger$Fre-Res Tokens refer to compact temporal-branch tokens, including dynamic P-Tokens, static tokens, and frequency-summary tokens. Raw spatial anchors are excluded to isolate the temporal-branch budget; in this setting, raw anchors consume $M \times 512$ tokens.}
\end{table}

\section{Additional Cross-Benchmark Ablation}
\label{app:cross_ablation}

Table~\ref{tab:ablation_cross} provides an additional cross-benchmark ablation to verify the complementary roles of the spatial and temporal tracks.
The main ablation in Table~\ref{tab:ablation} focuses on NExT-QA, where temporal and causal reasoning are central.
Here, we further evaluate the same branch-level variants on representative MVBench categories using the same Qwen2-VL-7B backbone.
This analysis is intended to study component behavior across temporal-heavy and spatial-heavy tasks, and is not directly comparable to the compact Qwen2-VL-2B setting in Table~\ref{tab:mvbench}.

The results show a clear division of labor.
The Temporal-Frequency Track performs strongly on NExT-QA, reaching 68.47\%, but its MVBench average drops to 55.12\%, with particularly low performance on Count and Pose.
In contrast, the Spatial Anchor Track obtains a stronger MVBench average of 64.83\%, indicating that high-fidelity spatial evidence is important for fine-grained visual reasoning.
The Dual-Track model with linear concatenation improves some spatial categories but remains weaker than the proposed absorber.
With the Spatial-Guided Absorber, Fre-Res achieves the best performance on both NExT-QA and MVBench average, suggesting that the absorber helps integrate temporal-frequency dynamics with spatial visual evidence.

\begin{table}[t]
\centering
\caption{\textbf{Cross-benchmark ablation of complementary track roles.}
Branch-level variants are evaluated on both NExT-QA and representative MVBench categories using the same Qwen2-VL-7B backbone.
The results indicate a clear division of labor: the temporal-frequency track performs strongly on causal reasoning (NExT-QA) but degrades on spatially sensitive tasks such as Count and Pose.
Naive concatenation suffers from semantic mismatch, whereas the Spatial-Guided Absorber effectively integrates both tracks across temporal and spatial reasoning tasks.}
\label{tab:ablation_cross}
\resizebox{\columnwidth}{!}{
\begin{tabular}{l c | c c c c}
\toprule
\textbf{Configuration} & \textbf{NExT-QA} 
& \textbf{MVBench Avg.} & \textbf{Action} & \textbf{Count} & \textbf{Pose} \\
\midrule
(A) Spatial Anchor Only 
& 65.59 & 64.83 & 63.41 & 56.78 & 51.24 \\
(B) Temporal-Frequency Only  
& 68.47 & 55.12 & 63.95 & 41.33 & 36.82 \\
(C) Dual-Track, no Absorber (Linear Concat)
& 66.42 & 63.76 & 65.02 & 55.41 & 49.65 \\
\rowcolor{blue!5} 
\textbf{(D) Dual-Track + Absorber (Ours)} 
& \textbf{70.08} & \textbf{67.92} & \textbf{68.14} & \textbf{61.27} & \textbf{52.86} \\
\bottomrule
\end{tabular}
}
\end{table}
\section{Algorithms}
\label{app:algorithms}

\subsection{Budget-Adaptive Fre-Res Compression}

\begin{algorithm}[t]
\caption{Budget-Adaptive Fre-Res Token Construction}
\label{alg:fre_res}
\begin{algorithmic}[1]
\Require Video frames $\{x_t\}_{t=1}^{T}$, visual encoder $\Phi$, target token budget $B$
\Require Anchor count $M$, raw-token budget per anchor $K_{\mathrm{raw}}$, temporal grouping scale $G$, maximum DCT coefficient budget $K_{\max}$
\Ensure Compressed visual-token sequence $H_{\mathrm{vis}}$
\Statex \textit{// Anchor selection combines uniform temporal coverage with optional event-aware insertion.}
\Statex \textit{// The temporal-frequency budget is computed after reserving spatial, summary, and static tokens.}
\Statex \textit{// Low-frequency coefficients are selected per pooled spatial position and filtered by residual energy.}

\State Encode frames into latent visual tokens: $F_t = \Phi(x_t)$
\State Select sparse anchor frames $\mathcal{A} = \{a_1,\ldots,a_M\}$ using uniform sampling and optional event-aware insertion:
\[
    \mathrm{cos\_dist}(F_t,F_{t-1}) > \tau .
\]

\State Initialize raw-anchor token set $H_{\mathrm{raw}} \leftarrow \emptyset$
\For{each anchor frame $a \in \mathcal{A}$}
    \State Partition $F_a$ into local $3 \times 3$ spatial blocks
    \State In each block, remove the token with the lowest energy
    \State Retain at most $K_{\mathrm{raw}}$ raw tokens from frame $a$
    \State Add retained raw-anchor tokens to $H_{\mathrm{raw}}$
\EndFor

\State Compute spatial budget $B_{\mathrm{spatial}} = M \cdot K_{\mathrm{raw}}$
\State Reserve summary-token and static-token budgets $B_{\mathrm{summary}}$ and $B_{\mathrm{static}}$
\State Compute temporal-frequency budget:
\[
    B_{\mathrm{freq}}
    =
    B
    -
    B_{\mathrm{spatial}}
    -
    B_{\mathrm{summary}}
    -
    B_{\mathrm{static}} .
\]
\State Determine the retained DCT coefficient budget:
\[
    K
    =
    \min
    \left(
    K_{\max},
    L_{\mathrm{group}},
    \left\lfloor
    \frac{B_{\mathrm{freq}}}
    {N_{\mathrm{group}} \cdot N_{\mathrm{pool}}}
    \right\rfloor
    \right),
\]
where $L_{\mathrm{group}}$ is the effective residual length of each temporal group, $N_{\mathrm{group}}$ is the number of temporal groups, and $N_{\mathrm{pool}}$ is the number of pooled spatial positions.

\State Initialize Fre-Res token set $H_{\mathrm{freq}} \leftarrow \emptyset$
\For{each temporal group between neighboring anchors}
    \For{each non-anchor frame $t$ in the group}
        \State Find the nearest previous anchor frame $a(t)$
        \State Compute latent residual $\Delta F_t = F_t - F_{a(t)}$
    \EndFor
    \State Stack residuals along the temporal axis
    \State Spatially pool residuals into $N_{\mathrm{pool}}$ positions
    \State Apply temporal 1D-DCT to the pooled residual trajectories
    \State Retain the first $K$ low-frequency coefficients per pooled spatial position
    \State Filter near-zero-energy residual positions so that the temporal branch satisfies $B_{\mathrm{freq}}$
    \State Construct temporal P-Tokens and frequency-summary tokens
    \State Add compressed Fre-Res tokens to $H_{\mathrm{freq}}$
\EndFor

\State Apply RawAdapter and type embeddings to $H_{\mathrm{raw}}$
\State Apply Fre-Res token adapters to $H_{\mathrm{freq}}$
\State Use the Spatial-Guided Absorber to align temporal-frequency tokens with spatial anchors
\State Fuse the two streams with branch gates
\State \Return $H_{\mathrm{vis}}$
\end{algorithmic}
\end{algorithm}

\subsection{Spatial-Guided Absorber}

\begin{algorithm}[t]
\caption{Spatial-Guided Absorption of Temporal-Frequency Tokens}
\label{alg:absorber}
\begin{algorithmic}[1]
\Require Spatial anchor tokens $H_I \in \mathbb{R}^{N_I \times d}$ with coordinates $C_I$
\Require Temporal-frequency P-Tokens $H_P \in \mathbb{R}^{N_P \times d}$ with coordinates $C_P$
\Require Local neighborhood radius $r$
\Ensure Motion-aware anchor tokens $H_{\mathrm{dyn}}$

\State Construct a spatial mask $M \in \mathbb{R}^{N_I \times N_P}$
\For{each anchor token $i$ and P-Token $j$}
    \If{$\mathrm{dist}(C_I^i, C_P^j) \le r$}
        \State $M_{ij} \leftarrow 0$
    \Else
        \State $M_{ij} \leftarrow -\infty$
    \EndIf
\EndFor

\State Compute queries, keys, and values:
\[
Q = H_I W_Q, \quad K = H_P W_K, \quad V = H_P W_V.
\]
\State Apply local masked cross-attention:
\[
A = \mathrm{softmax}\left(\frac{QK^\top}{\sqrt{d}} + M\right).
\]
\State Aggregate temporal-frequency evidence:
\[
\tilde{H}_I = A V.
\]
\State Produce motion-aware anchors:
\[
H_{\mathrm{dyn}} = \mathrm{LayerNorm}(H_I + \tilde{H}_I).
\]
\State \Return $H_{\mathrm{dyn}}$
\end{algorithmic}
\end{algorithm}

\section{Impact on Model Output}
\label{app:impact_output}

Fre-Res changes the visual evidence provided to the LLM.
Instead of passing dense frame-wise visual tokens, it provides sparse spatial anchors and compact temporal-frequency residuals.
This can affect model outputs in both beneficial and failure-prone ways.

\textbf{Improved access to temporal evidence.}
By explicitly encoding inter-frame residual dynamics, Fre-Res can make short causal transitions more accessible to the LLM.
This is reflected in our NExT-QA ablations, where the temporal-frequency track alone performs strongly on causal reasoning.
Qualitatively, Fre-Res can recover transient interactions that may be missed by sparse frame sampling.

\textbf{Reduced static redundancy.}
Dense videos often contain many redundant background tokens.
Fre-Res reduces this redundancy and allows the model to focus on anchors and residual changes.
This can improve efficiency and reduce the burden of attending over many near-duplicate visual tokens.

\textbf{Potential loss of fine details.}
Compression may remove or weaken subtle visual cues.
Pose, small objects, text, fine-grained attributes, and rare local interactions can be sensitive to the chosen anchor count and spatial token-retention ratio.
When these details are crucial, more conservative budgets or query-aware anchor allocation may be required.

\textbf{Output shifts under compression.}
Because the compressed visual sequence is not identical to the full-token input, the model may produce different answers from the vanilla model.
Such output shifts are not necessarily errors: in some cases, compressed residual-frequency evidence may emphasize motion cues more clearly.
However, in ambiguous or detail-heavy examples, compression may also introduce incorrect answers.
For this reason, we report both accuracy and visual-token counts, and we analyze failure modes across task categories.

\textbf{Recommended usage.}
Fre-Res is best suited for efficient video QA and long-video reasoning where temporal coverage is important and full-token inference is expensive.
For safety-critical settings or tasks requiring precise local measurements, compressed outputs should be verified with higher-fidelity visual evidence or human review.

\section{Additional Qualitative Analysis}
\label{app:qualitative}

\begin{figure*}[t]
    \centering
    \includegraphics[width=\textwidth]{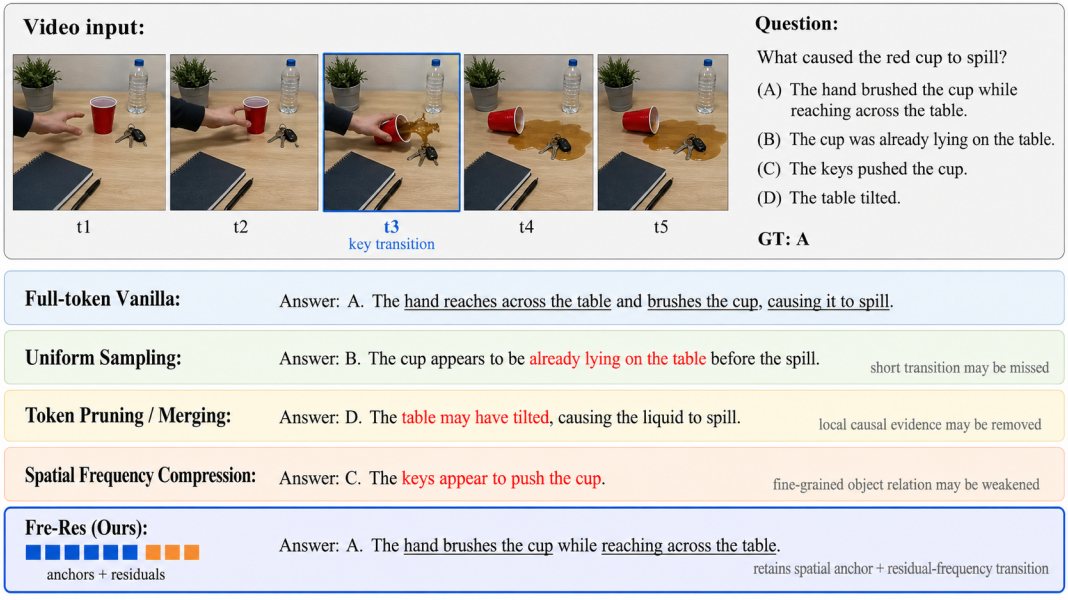}
    \caption{\textbf{Qualitative comparison on causal video reasoning.}
This example requires recognizing a short interaction between the hand and the red cup.
Different compression strategies preserve different evidence: sparse sampling may miss the transient frame, token pruning or merging may remove local interaction cues, and spatial frequency compression may weaken fine-grained object relations.
Fre-Res retains spatial anchors together with residual-frequency evidence, leading to the correct causal answer in this case.}
\label{fig:qualitative}
\end{figure*}

Figure~\ref{fig:qualitative} shows a qualitative causal reasoning example.
The question requires recognizing a short interaction between the hand and the red cup.
Sparse sampling may miss the transient interaction, token pruning or merging may remove local interaction cues, and frame-wise spatial frequency compression may weaken fine-grained object relations.
Fre-Res retains sparse spatial anchors together with residual-frequency evidence, enabling the model to recover the correct causal answer in this case.

\end{document}